\documentclass[sigconf]{acmart}
\usepackage{booktabs} 

\makeatletter
\renewcommand\@formatdoi[1]{\ignorespaces}
\makeatother

\setcopyright{rightsretained}

\acmConference[ECCV'18-EPIC]{3\textsuperscript{rd} International Egocentric, Perception, Interaction and Computing (EPIC) Workshop}{September 2018}{Munich, Germany}
\acmYear{2018}
\copyrightyear{2018}

\AtBeginDocument{\AtBeginShipoutNext{\AtBeginShipoutDiscard}}

\begin{document}
\title{Towards an Embodied Semantic Fovea: Semantic 3D scene reconstruction from ego-centric eye-tracker videos}

\author{Mickey Li}
\affiliation{%
  \institution{Brain \& Behaviour Lab,\\ Dept. of Computing, \\Imperial College London}
}

\author{Noyan Songur}
\affiliation{%
  \institution{Brain \& Behaviour Lab,\\ Dept. of Bioengineering,\\ Imperial College London}
}

\author{Pavel Orlov}
\orcid{0000-0001-6256-8134}
\affiliation{%
  \institution{Brain \& Behaviour Lab,\\ Dept. of Computing, \\Imperial College London}
}

\author{Stefan Leutenegger}
\affiliation{%
  \institution{Smart Robotics Lan\\Dept. of Computing,\\ Imperial College London}
}
\email{s.leutenegger@imperial.ac.uk}

\author{A. Aldo Faisal}\
\orcid{0000-0003-0813-7207}
\affiliation{
  \institution{Brain \& Behaviour Lab,\\ Dept. of Computing \& \\Dept. of Bioengineering,\\ Imperial College London}
 }
\email{a.faisal@imperial.ac.uk}

\renewcommand{\shortauthors}{Li et Faisal}

\begin{abstract}
Incorporating the physical environment is essential for a complete understanding of human behavior in unconstrained every-day tasks. This is especially important in ego-centric tasks where obtaining 3 dimensional information is both limiting and challenging with the current 2D video analysis methods proving insufficient. Here we demonstrate a proof-of-concept system which provides real-time 3D mapping and semantic labeling of the local environment from an ego-centric RGB-D video-stream with 3D gaze point estimation from head mounted eye tracking glasses. We augment existing work in Semantic Simultaneous Localization And Mapping (Semantic SLAM) with collected gaze vectors. Our system can then find and track objects both inside and outside the user field-of-view in 3D from multiple perspectives with reasonable accuracy. We validate our concept by producing a semantic map from images of the NYUv2 dataset while simultaneously estimating gaze position and gaze classes from recorded gaze data of the dataset images.
\end{abstract}





\maketitle

\section{Introduction}
Human behavior is the result of our perception and actions combined with our  goals and the physical environment within which they are embedded. In particular, human behavior is not isolated from the world, and our visual perception, although mostly studied in impoverished lab settings, is highly dynamic and drives as much as it is driven by our movements through scene.  Therefore, to study human behavior in the wild, we have to take into analyse the physical world as experienced through a subject's natural movement behavior.

In studies of human activity and human behavior, 2D ego-centric videos are typically used to understand the persons visual input \citep{Land1999TheLiving,Carrasco2012ExploitingMovements}. However, 2D images do not give us a complete representation of the real world which is three dimensional.  Moreover, eye-movements play a proactive role in human behavior activity and therefore, are increasingly studied \citep{Ballard1992Hand-eyeTasks,Hayhoe2003VisualTask,Rothkopf2007TaskLook}. Typically eye-movements are mostly monitored in form of an estimated gaze point mapped onto a 2D image, and we recently showed high-accuracy 3D gaze tracking in high-accuracy binocular eye-tracking \citep{Abbott2012Ultra-low-costInterfaces}. Although we can reconstruct the 3D gaze location, we would still need to know what object we are looking it. We know that the eye's fovea, the single spot of high resolution of the human eye, tracks the objects of our overt visual attention. Current solutions to measure human behavior reconstruct the fovea from an eye tracker mounted scene camera and use convolutional neural networks for real-time ego-centric object labeling from eye tracking camera feeds, the so called Semantic Fovea \cite{Auepanwiriyakul2018SemanticContext}, which tells us what object is currently looked at by the user. However, it still remains unclear how this object is related to other objects in the scene, where all these objects are located and, as we move through the scene freely, we would like to know we are looking at the same object again (e.g. a book shelf from two different sides) or different entities of the same object category. Therefore, we present here our first steps towards an \textbf{embodied} Semantic Fovea.

\begin{figure*}
\includegraphics[width=\textwidth]{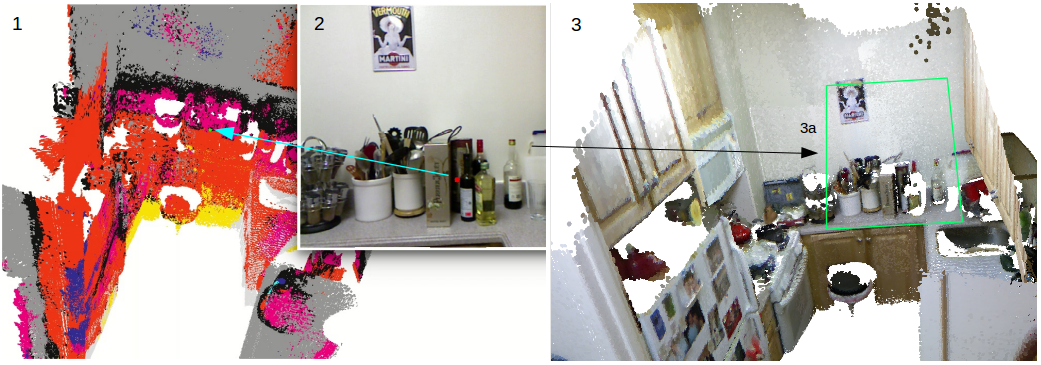}
\caption{Embodied Semantic Fovea in action. The SLAM system takes a sequence of images and generates a 3D surfel map. Two similar views into the same map are shown in (1) and (3). (3) is the real color reconstruction from the image sequence and (1) shows the class-label reconstruction where each color represents a different class type, e.g. pink:furniture, orange:object. (2) is one particular frame in the construction which corresponds to the area shown in (3a) and the equivalent area in (1). In (2) the red dot denotes the recorded gaze point for that frame. This gaze point 3D location is estimated and placed back into (1) as described by the green surfels, as indicated by the arrow. We therefore also know what object the gaze point corresponds to.}
\label{fig:esf1}
\end{figure*}

In 2D video object tracking, we can track an object when it is on the frame, but we lose it immediately when it is not. Thus, in subsequent interactions, we do not know whether it is the same object as before. In addition 2D images that are often used for analysis are likely to contain perspective and geometric issues which affects the object properties, for instance, the size of two identical cups can be skewed by perspective. Previously, external motion tracking systems with markers on both the objects and subjects\cite{HajiFathaliyan2018ExploitingCollaboration} where used for both issues, and wrist-mounted video cameras\cite{Carrasco2012ExploitingMovements} for the former. However, motion tracking systems do not work well in the wild because of the physical limitations, and wrist-mounted video cameras also suffer from the same 2D object tracking issues. 

We have therefore built a system that allows a user to interact with physical objects with their gaze during everyday activities. With a SLAM algorithm our system builds a 3D representation of the surroundings upon which the user's gaze is placed. The 3D representation knows the object's pose, texture and type. This allows for persistent object tracking as a user can look away and back to an object as the system registers it as the same object. This gives us the ability to monitor users' gaze in real space so we can generate 3D tracking of dynamic gaze locations which can offer greater insights into human behavior research.

\section{Method}
Our system is comprised of a real-time semantically labeled simultaneous localization and mapping (SLAM) module known as SemanticFusion\cite{McCormac2016SemanticFusion:ArXiv}, and a 3D gaze estimation method\cite{Abbott2012Ultra-low-costInterfaces}. The role of the semantic SLAM system is to take the live RGB-D egocentric frames from a camera mounted on the user's forehead to build up a globally consistent, semantically labeled 3D map of the user's surrounding area. The labeling is generated by a CNN which takes the 2D frames to return a set of per-pixel class probabilities. Our system updates the original SemanticFusion CNN with a state-of-the-art network. Finally, the gaze estimation takes the 2D gaze locations from the eye tracking glasses and estimates their 3D location within this semantically labeled map.

\begin{figure*}
\includegraphics[width=\textwidth]{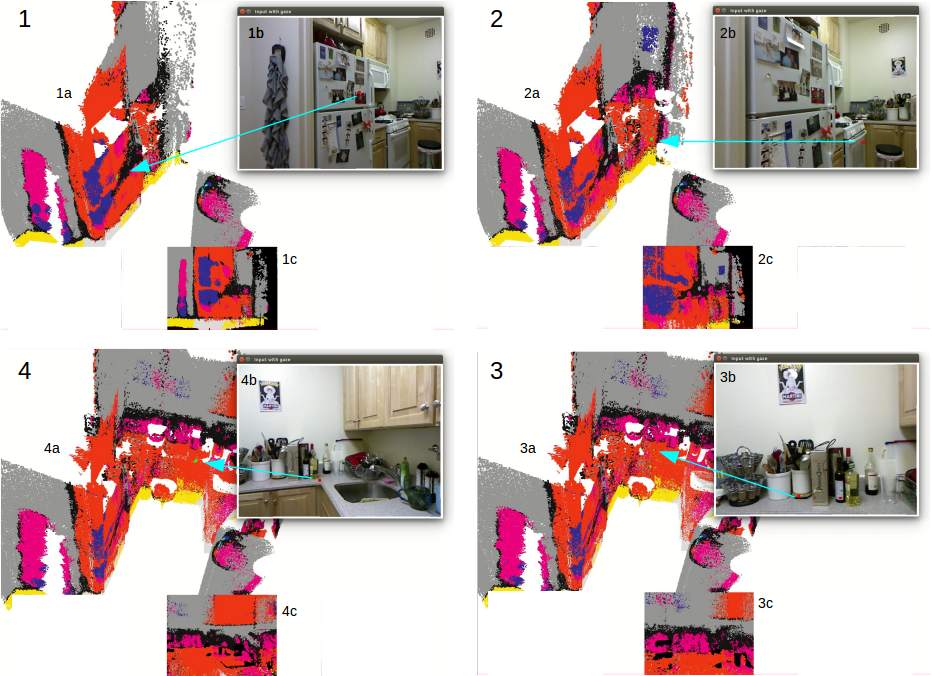}
\caption{A selection of frames from Embodied Semantic Fovia. Similar to Figure \ref{fig:esf1}, the (a) surfaces show the SLAM reconstruction in action from (1 to 4). The colours on the surfaces again represent the class of any given surfel. (c) shows a view of the surface (a) from the estimated camera pose of input image (b). In each (b) we see the recorded gaze location which is projected to a 3D surfel in (a) which is coloured green as indicated by the arrows. Taking reference from the RGB reconstruction of figure \ref{fig:esf1}, we see that the estimation is reasonably accurate.}
\label{fig:esf2}
\end{figure*}

\subsection{Semantic SLAM mapping}
We use SemanticFusion\cite{McCormac2016SemanticFusion:ArXiv} as basis for our semantic SLAM component. SemanticFusion is composed  of the ElasticFusion\cite{Whelan2016ElasticFusion:Estimation} SLAM system which builds a surface element (surfel) map of the environment and a convolutional neural network which provides per pixel labeling for each 2D input frame which is then projected to the surfel map. For each arriving image frame $k$, ElasticFusion tracks the pose of the camera by aligning RGB frames using the \textit{iterative closest points} (ICP) algorithm on the surfel map. A \textit{surfel} is a 3D disc in space with properties such as a surface normal, color, and for semantic mappings, a class label. To update the surfel map, the given camera pose is used to map pixels of the camera image to surfels in the camera's field-of-view. Comparing the map to the new image, new surfels are initialized using the collected depth value for pixels which have no surfel mapping. Otherwise, the existing information is updated by combining them with the new evidence from the current frame. 
Figure \ref{fig:esf1} demonstrates a map created by this system. Note that the camera pose tracking is equivalent to tracking the position of the user's head and gaze direction within the mapped environment.

The use of surfels allows SemanticFusion to provide a mapping from class probability to surfel and for those mappings to be `carried along' when the map updates or in loop closures. The semantic class probabilities are generated using a Fully Convolutional Network (FCN)\cite{LongFullySegmentation} which is a convolutional network that was trained to outputs a per-pixel class probability distribution. With the known camera pose, a pixel is assigned to a surfel and a \textit{Bayesian update scheme} is used to refine the class probabilities for that surfel. SemanticFusion operates at real-time frame-rates on VGA resolutions and makes use of the Microsoft Kinect depth camera.  Our system component SemanticFusion by allowing the use of the smaller, more lightweight \textit{Intel RealSense} depth camera (and its data formats) which can be mounted on a subject's forehead. This makes this system ideal for accurate real-time semantic reconstruction of everyday scenarios.

\begin{figure}[t]
\includegraphics[width=\columnwidth]{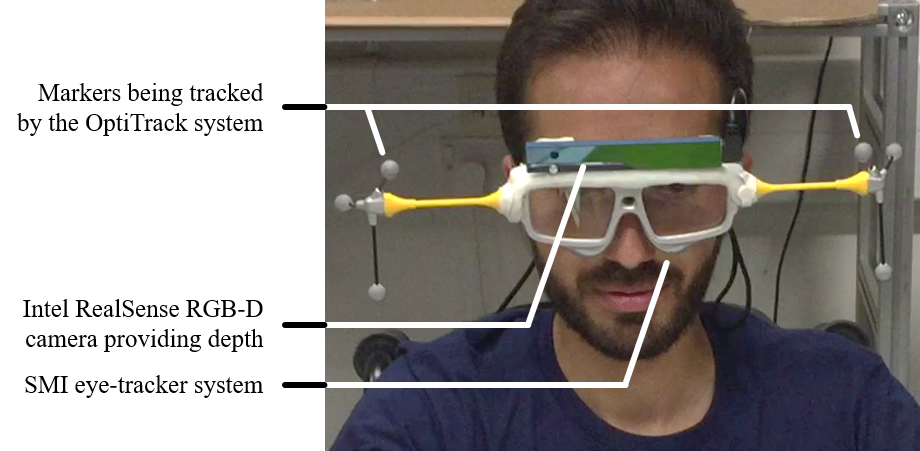}
\caption{The eye tracking set-up consists of wearable binocular eye tracking glasses and a RGB-D camera mounted on top. The 3D optical motion tracking markers are only used to validate our semantic reconstruction against ground truth data.}
\label{fig:setup}
\end{figure}

\subsection{Deep Semantic Labeling}
For the scenarios in which we wish to use our system, a more powerful labeling method is required as it would allow greater reliability and better accuracy in the detection of a greater number of objects which is necessary for a more robust system.
In this work we use state-of-the art network for semantic segmentation known as the DeepLab (V3+) network developed by Papandreau et al.\citep{PapandreouWeakly-andSegmentation, ChenEncoder-DecoderSegmentation} which makes use of \textit{atruos spatial pyramid pooling} to robustly segment objects at multiple scales. This network outperforms the original FCN network of SemanticFusion in accuracy and in the detection robustness for larger numbers of classes. A particular reason for choosing this network is in observing the networks performance on the \textit{ADE20k Dataset} which contains 22,210 segmented images containing 151 different classes (one of which is the background). The performance with the dataset proves that this network is still robust with larger numbers of classes where other networks struggle. The downside of this particular network is there is a speed-accuracy trade off with the DeepLab network running at most at 5 frames per second(fps). We based our model on the author's implementation of DeepLab in \textit{TensorFlow}, along with pre-trained weights after training on the ADE20k dataset, and integrated it with SemanticFusion for a more flexible semantic segmentation network. 

\subsection{3D Gaze Estimation}
With the labeled 3D map built by SemanticFusion, it is now possible to easily estimate 3D gaze locations. By monitoring the location of the user's pupils, the eye tracking glasses return a $(x, y)$ pixel position within the ego-centric image which represents the current gaze. Once the gaze location is found with respect the image, SemanticFusion then provides us a mapping between the gaze pixel and its corresponding surfel in the map. This is because when the map is built, we require the 2D projection of our 3D surfel map onto the camera where every pixel of this 2D projection is a corresponding surfel (if the surfel exists). The results of this technique are shown in Figure \ref{fig:esf2} for a sequence of pre-captured images within the NYUv2 dataset and a simulated set of gaze points. 

For real-time use, the setup shown in Figure \ref{fig:setup} is the platform that is developed for the analysis of embodied semantic fovea in the wild. As before the gaze informatics are provided by the eye tracking glasses, but the rgb and depth images are captured using the \textit{Intel RealSense} RGB-D camera mounted on top of the glasses. The gaze is then overlaid on the RGB-D field-of-view using projective transformation. The full setup will enable the estimation of the 3D gaze point along with the semantics of the visual attention which addresses many of the challenges with 2D gaze informatics tools.

\section{Conclusions}
The use of ego-centric SLAM allows us to create a detailed 3D representation of the real-world which is more complete than existing 2D methods. Generating a 3D representation also allows easy estimation of 3D gaze points and as a result we do not lose any of the semantics of the interactions. 

Furthermore, the system allows tracking of a single object from multiple perspectives as a single complete map is generated. Similarly, using a single map also resolves the perspective and geometric issues stemming from 2D systems. On top of this, the use of semantic labeling to label the map allows the discovery of specific objects within the map. Finally, the use of SLAM with no reliance on external cameras also removes the limitations of currently available motion tracking systems. 

We foresee that the primary use of the system is for researchers who have a focus on the embodied nature of the human behavior. For example, this system is used in the eNHANCE project where a more complete understanding of human behavior is required to build a gaze-contingent robotic assistant system. However, there are many other applications for this system in many fields, including complex multi-users (or robots) interaction, AR, VR, and robotic teleoperation.

\begin{acks}
The authors would like to thank Chaiyawan Auepanwiriyakul and Alex 
Harston for technical discussions. This research was supported by the eNHANCE Project (http://www.enhance-motion.eu) under the European  Union's Horizon 2020 research and innovation programme grant agreement No. 644000.
\end{acks}

\bibliographystyle{ACM-Reference-Format}
\bibliography{Mendeley}


\begin{thebibliography}{13}


\ifx \showCODEN    \undefined \def \showCODEN     #1{\unskip}     \fi
\ifx \showDOI      \undefined \def \showDOI       #1{#1}\fi
\ifx \showISBNx    \undefined \def \showISBNx     #1{\unskip}     \fi
\ifx \showISBNxiii \undefined \def \showISBNxiii  #1{\unskip}     \fi
\ifx \showISSN     \undefined \def \showISSN      #1{\unskip}     \fi
\ifx \showLCCN     \undefined \def \showLCCN      #1{\unskip}     \fi
\ifx \shownote     \undefined \def \shownote      #1{#1}          \fi
\ifx \showarticletitle \undefined \def \showarticletitle #1{#1}   \fi
\ifx \showURL      \undefined \def \showURL       {\relax}        \fi
\providecommand\bibfield[2]{#2}
\providecommand\bibinfo[2]{#2}
\providecommand\natexlab[1]{#1}
\providecommand\showeprint[2][]{arXiv:#2}

\bibitem[\protect\citeauthoryear{Abbott and Faisal}{Abbott and Faisal}{2012}]%
        {Abbott2012Ultra-low-costInterfaces}
\bibfield{author}{\bibinfo{person}{William~Welby Abbott} {and}
  \bibinfo{person}{Aldo~Ahmed Faisal}.} \bibinfo{year}{2012}\natexlab{}.
\newblock \showarticletitle{{Ultra-low-cost 3D gaze estimation: an intuitive
  high information throughput compliment to direct brain–machine
  interfaces}}.
\newblock \bibinfo{journal}{\emph{Journal of neural engineering}}
  \bibinfo{volume}{9}, \bibinfo{number}{4} (\bibinfo{year}{2012}),
  \bibinfo{pages}{046016}.
\newblock
\showISSN{1741-2552}


\bibitem[\protect\citeauthoryear{Auepanwiriyakul, Harston, Orlov, Shafti, and
  Faisal}{Auepanwiriyakul et~al\mbox{.}}{2018}]%
        {Auepanwiriyakul2018SemanticContext}
\bibfield{author}{\bibinfo{person}{Chaiyawan Auepanwiriyakul},
  \bibinfo{person}{Alex Harston}, \bibinfo{person}{Pavel Orlov},
  \bibinfo{person}{Ali Shafti}, {and} \bibinfo{person}{A~Aldo Faisal}.}
  \bibinfo{year}{2018}\natexlab{}.
\newblock \showarticletitle{{Semantic Fovea: Real-time annotation of
  ego-centric videos with gaze context}}.
\newblock   \bibinfo{volume}{18} (\bibinfo{year}{2018}).
\newblock
\urldef\tempurl%
\url{https://doi.org/10.1145/3204493.3208349}
\showDOI{\tempurl}


\bibitem[\protect\citeauthoryear{Ballard, Hayhoe, Li, and Whitehead}{Ballard
  et~al\mbox{.}}{1992}]%
        {Ballard1992Hand-eyeTasks}
\bibfield{author}{\bibinfo{person}{D~H Ballard}, \bibinfo{person}{M~M Hayhoe},
  \bibinfo{person}{F Li}, {and} \bibinfo{person}{S~D Whitehead}.}
  \bibinfo{year}{1992}\natexlab{}.
\newblock \showarticletitle{{Hand-eye coordination during sequential tasks}}.
\newblock \bibinfo{journal}{\emph{Philos. Trans. R. Soc. Lond. B Biol. Sci.}}
  \bibinfo{volume}{337}, \bibinfo{number}{1281} (\bibinfo{year}{1992}),
  \bibinfo{pages}{331--8}.
\newblock
\showISSN{0962-8436}
\urldef\tempurl%
\url{https://doi.org/10.1098/rstb.1992.0111}
\showDOI{\tempurl}


\bibitem[\protect\citeauthoryear{Carrasco and Clady}{Carrasco and
  Clady}{2012}]%
        {Carrasco2012ExploitingMovements}
\bibfield{author}{\bibinfo{person}{Miguel Carrasco} {and}
  \bibinfo{person}{Xavier Clady}.} \bibinfo{year}{2012}\natexlab{}.
\newblock \showarticletitle{{Exploiting eye–hand coordination to detect
  grasping movements}}.
\newblock \bibinfo{journal}{\emph{Image and Vision Computing}}
  \bibinfo{volume}{30}, \bibinfo{number}{11} (\bibinfo{date}{11}
  \bibinfo{year}{2012}), \bibinfo{pages}{860--874}.
\newblock
\showISSN{0262-8856}
\urldef\tempurl%
\url{https://doi.org/10.1016/J.IMAVIS.2012.07.001}
\showDOI{\tempurl}


\bibitem[\protect\citeauthoryear{Chen, Zhu, Papandreou, Schroff, and Adam}{Chen
  et~al\mbox{.}}{[n. d.]}]%
        {ChenEncoder-DecoderSegmentation}
\bibfield{author}{\bibinfo{person}{Liang-Chieh Chen}, \bibinfo{person}{Yukun
  Zhu}, \bibinfo{person}{George Papandreou}, \bibinfo{person}{Florian Schroff},
  {and} \bibinfo{person}{Hartwig Adam}.} \bibinfo{year}{[n. d.]}\natexlab{}.
\newblock \showarticletitle{{Encoder-Decoder with Atrous Separable Convolution
  for Semantic Image Segmentation}}.
\newblock  (\bibinfo{year}{[n. d.]}).
\newblock
\urldef\tempurl%
\url{https://github.com/tensorflow/}
\showURL{%
\tempurl}


\bibitem[\protect\citeauthoryear{Haji~Fathaliyan, Wang, and
  Santos}{Haji~Fathaliyan et~al\mbox{.}}{2018}]%
        {HajiFathaliyan2018ExploitingCollaboration}
\bibfield{author}{\bibinfo{person}{Alireza Haji~Fathaliyan},
  \bibinfo{person}{Xiaoyu Wang}, {and} \bibinfo{person}{Veronica~J. Santos}.}
  \bibinfo{year}{2018}\natexlab{}.
\newblock \showarticletitle{{Exploiting Three-Dimensional Gaze Tracking for
  Action Recognition During Bimanual Manipulation to Enhance Human–Robot
  Collaboration}}.
\newblock \bibinfo{journal}{\emph{Frontiers in Robotics and AI}}
  \bibinfo{volume}{5} (\bibinfo{date}{4} \bibinfo{year}{2018}),
  \bibinfo{pages}{25}.
\newblock
\showISSN{2296-9144}
\urldef\tempurl%
\url{https://doi.org/10.3389/frobt.2018.00025}
\showDOI{\tempurl}


\bibitem[\protect\citeauthoryear{Hayhoe, Shrivastava, Mruczek, and Pelz}{Hayhoe
  et~al\mbox{.}}{2003}]%
        {Hayhoe2003VisualTask}
\bibfield{author}{\bibinfo{person}{Mary~M. Hayhoe}, \bibinfo{person}{Anurag
  Shrivastava}, \bibinfo{person}{Ryan Mruczek}, {and} \bibinfo{person}{Jeff~B.
  Pelz}.} \bibinfo{year}{2003}\natexlab{}.
\newblock \showarticletitle{{Visual memory and motor planning in a natural
  task}}.
\newblock \bibinfo{journal}{\emph{Journal of Vision}} \bibinfo{volume}{3},
  \bibinfo{number}{1} (\bibinfo{date}{2} \bibinfo{year}{2003}),
  \bibinfo{pages}{6}.
\newblock
\showISSN{1534-7362}
\urldef\tempurl%
\url{https://doi.org/10.1167/3.1.6}
\showDOI{\tempurl}


\bibitem[\protect\citeauthoryear{Land, Mennie, and Rusted}{Land
  et~al\mbox{.}}{1999}]%
        {Land1999TheLiving}
\bibfield{author}{\bibinfo{person}{Michael Land}, \bibinfo{person}{Neil
  Mennie}, {and} \bibinfo{person}{Jennifer Rusted}.}
  \bibinfo{year}{1999}\natexlab{}.
\newblock \showarticletitle{{The Roles of Vision and Eye Movements in the
  Control of Activities of Daily Living}}.
\newblock \bibinfo{journal}{\emph{Perception}} \bibinfo{volume}{28},
  \bibinfo{number}{11} (\bibinfo{date}{11} \bibinfo{year}{1999}),
  \bibinfo{pages}{1311--1328}.
\newblock
\showISSN{0301-0066}
\urldef\tempurl%
\url{https://doi.org/10.1068/p2935}
\showDOI{\tempurl}


\bibitem[\protect\citeauthoryear{Long, Shelhamer, and Darrell}{Long
  et~al\mbox{.}}{[n. d.]}]%
        {LongFullySegmentation}
\bibfield{author}{\bibinfo{person}{Jonathan Long}, \bibinfo{person}{Evan
  Shelhamer}, {and} \bibinfo{person}{Trevor Darrell}.} \bibinfo{year}{[n.
  d.]}\natexlab{}.
\newblock \showarticletitle{{Fully Convolutional Networks for Semantic
  Segmentation}}.
\newblock  (\bibinfo{year}{[n. d.]}).
\newblock
\urldef\tempurl%
\url{https://people.eecs.berkeley.edu/~jonlong/long_shelhamer_fcn.pdf}
\showURL{%
\tempurl}


\bibitem[\protect\citeauthoryear{McCormac, Handa, Davison, and
  Leutenegger}{McCormac et~al\mbox{.}}{2016}]%
        {McCormac2016SemanticFusion:ArXiv}
\bibfield{author}{\bibinfo{person}{J McCormac}, \bibinfo{person}{A Handa},
  \bibinfo{person}{A Davison}, {and} \bibinfo{person}{S Leutenegger}.}
  \bibinfo{year}{2016}\natexlab{}.
\newblock \showarticletitle{{SemanticFusion: dense 3D semantic mapping with
  convolutional neural networks arXiv]}}.
\newblock \bibinfo{journal}{\emph{arXiv}} (\bibinfo{year}{2016}),
  \bibinfo{pages}{7 pp.}
\newblock
\urldef\tempurl%
\url{http://arxiv.org/abs/1609.05130 http://arxiv.org/abs/1609.05130}
\showURL{%
\tempurl}


\bibitem[\protect\citeauthoryear{Papandreou, Chen, Murphy, and
  Yuille~Ucla}{Papandreou et~al\mbox{.}}{[n. d.]}]%
        {PapandreouWeakly-andSegmentation}
\bibfield{author}{\bibinfo{person}{George Papandreou},
  \bibinfo{person}{Liang-Chieh Chen}, \bibinfo{person}{Kevin~P Murphy}, {and}
  \bibinfo{person}{Alan~L Yuille~Ucla}.} \bibinfo{year}{[n. d.]}\natexlab{}.
\newblock \showarticletitle{{Weakly-and Semi-Supervised Learning of a Deep
  Convolutional Network for Semantic Image Segmentation}}.
\newblock  (\bibinfo{year}{[n. d.]}).
\newblock
\urldef\tempurl%
\url{http://openaccess.thecvf.com/content_iccv_2015/papers/Papandreou_Weakly-_and_Semi-Supervised_ICCV_2015_paper.pdf}
\showURL{%
\tempurl}


\bibitem[\protect\citeauthoryear{Rothkopf, Ballard, and Hayhoe}{Rothkopf
  et~al\mbox{.}}{2007}]%
        {Rothkopf2007TaskLook}
\bibfield{author}{\bibinfo{person}{Constantin~A Rothkopf},
  \bibinfo{person}{Dana~H Ballard}, {and} \bibinfo{person}{Mary~M Hayhoe}.}
  \bibinfo{year}{2007}\natexlab{}.
\newblock \showarticletitle{{Task and context determine where you look}}.
\newblock \bibinfo{journal}{\emph{J. Vis.}} \bibinfo{volume}{7},
  \bibinfo{number}{14} (\bibinfo{year}{2007}), \bibinfo{pages}{16}.
\newblock
\showISSN{1534-7362}
\urldef\tempurl%
\url{https://doi.org/10.1167/7.14.16}
\showDOI{\tempurl}


\bibitem[\protect\citeauthoryear{Whelan, Renato, Glocker, Andrew, and
  Leutenegger}{Whelan et~al\mbox{.}}{2016}]%
        {Whelan2016ElasticFusion:Estimation}
\bibfield{author}{\bibinfo{person}{Thomas Whelan}, \bibinfo{person}{F~Salas
  Renato}, \bibinfo{person}{Ben Glocker}, \bibinfo{person}{J~Davison Andrew},
  {and} \bibinfo{person}{Stefan Leutenegger}.} \bibinfo{year}{2016}\natexlab{}.
\newblock \showarticletitle{{ElasticFusion: Real-time dense SLAM and light
  source estimation}}.
\newblock \bibinfo{journal}{\emph{The International Journal of Robotics
  Research}} \bibinfo{volume}{35}, \bibinfo{number}{14} (\bibinfo{year}{2016}),
  \bibinfo{pages}{1697--1716}.
\newblock
\showISSN{0278-3649}
\urldef\tempurl%
\url{https://doi.org/10.1177/0278364916669237}
\showDOI{\tempurl}


\end{thebibliography}

\end{document}